\documentclass[10pt,twocolumn,letterpaper]{article}

\usepackage{iccv}
\usepackage{times}
\usepackage{epsfig}
\usepackage{graphicx}
\usepackage{amsmath}
\usepackage{amssymb}
\usepackage{makecell}
\usepackage{xspace}
\usepackage{comment}
\usepackage{caption}
\usepackage{subcaption}
\usepackage[accsupp]{axessibility}  % Improves PDF readability for those with disabilities.

\newcommand{\methodone}{Ours (-real data)\xspace} %
\newcommand{\methodtwo}{Ours (-optim. phys.)\xspace} %
\newcommand{\lossnameone}{2D perceptual loss (rendering)\xspace} %
\usepackage[pagebackref=true,breaklinks=true,letterpaper=true,colorlinks,bookmarks=false]{hyperref}

\usepackage[capitalize]{cleveref}
\crefname{section}{Sec.}{Secs.}
\Crefname{section}{Section}{Sections}
\Crefname{table}{Table}{Tables}
\crefname{table}{Tab.}{Tabs.}

\iccvfinalcopy % *** Uncomment this line for the final submission

\def\iccvPaperID{8923} % *** Enter the ICCV Paper ID here

\ificcvfinal\fi

\begin{document}

\definecolor{todocolor}{RGB}{255,0,00}
\newcommand\TODO[1] {\PackageWarning{}{Unprocessed todo}\emph{\textcolor{todocolor}{TODO: #1}}}

\renewcommand{\paragraph}[1]{\vspace{.125cm}\noindent\textbf{#1}}

%%%%%%%%% TITLE
%\title{Clothing Simulation and Digital Human Reconstruction}
\title{CaPhy: Capturing Physical Properties for Animatable Human Avatars}
\author{Zhaoqi Su\\
Tsinghua University\\
Beijing, China\\
{\tt\small suzq13@tsinghua.org.cn}
\and
Liangxiao Hu\\
Harbin Institute of Technology\\
Weihai, Shandong, China\\
{\tt\small lx.hu@hit.edu.cn}
\and
Siyou Lin\\
Tsinghua University\\
Beijing, China\\
{\tt\small linsy21@mails.tsinghua.edu.cn}
\and
Hongwen Zhang\\
Tsinghua University\\
Beijing, China\\
{\tt\small zhanghongwen@tsinghua.edu.cn}
\and
Shengping Zhang\\
Harbin Institute of Technology\\
Weihai, Shandong, China\\
{\tt\small s.zhang@hit.edu.cn}
\and
Justus Thies\\
Max Planck Institute for Intelligent Systems\\
T{\"u}bingen, Germany\\
{\tt\small justus.thies@tuebingen.mpg.de}
\and
Yebin Liu\\
Tsinghua University\\
Beijing, China\\
{\tt\small liuyebin@mail.tsinghua.edu.cn}
}

%\maketitle

\twocolumn[{%
	\renewcommand\twocolumn[1][]{#1}%
	\maketitle
	\begin{center}
		\centerline{
                \includegraphics[width=\textwidth]{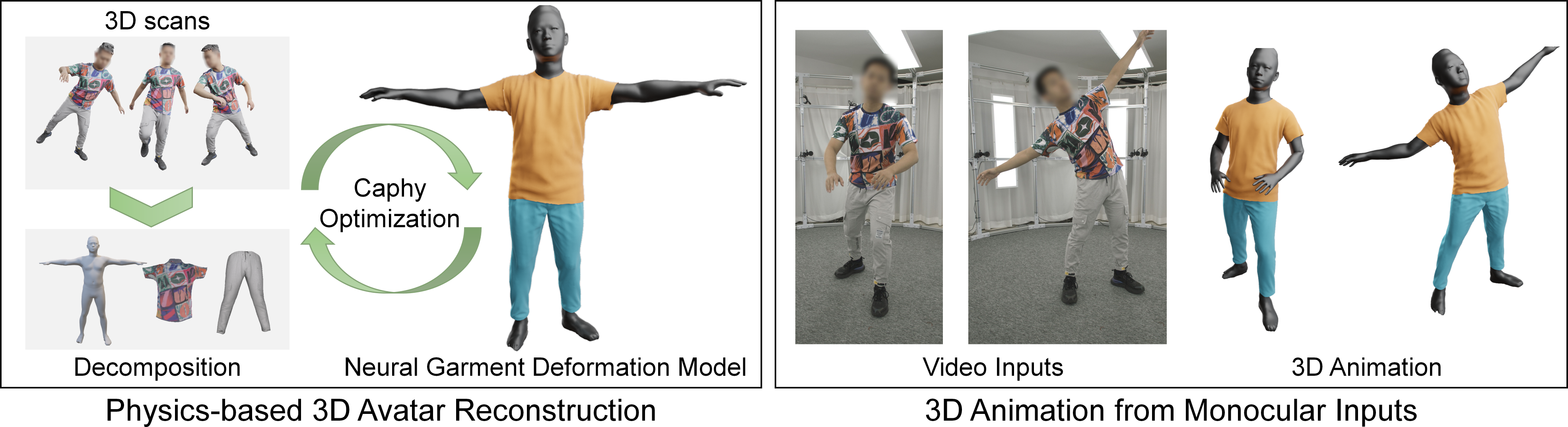}
		}
        \captionof{figure}{
        We propose CaPhy, a novel method for reconstructing physics-based 3D human avatars from 3D scans. We optimize a garment deformation model that is based on a deep neural network using real-world observations to capture the physical properties of the clothing. The neural garment deformation model is conditioned on the pose of the underlying body model, allowing us to repose the avatar with a realistic synthesis of cloth wrinkles. Specifically, the reconstructed avatar can be controlled by monocular RGB input data which is used to estimate the body pose.
        }
        \label{fig:teaser}
	\end{center}
}]

% Remove page # from the first page of camera-ready.
\ificcvfinal\thispagestyle{empty}\fi

\begin{abstract}
We present CaPhy, a novel method for reconstructing animatable human avatars with realistic dynamic properties for clothing. 
Specifically, we aim for capturing the geometric and physical properties of the clothing from real observations.
This allows us to apply novel poses to the human avatar with physically correct deformations and wrinkles of the clothing.
To this end, we combine unsupervised training with physics-based losses and 3D-supervised training using scanned data to reconstruct a dynamic model of clothing that is physically realistic and conforms to the human scans.
We also optimize the physical parameters of the underlying physical model from the scans by introducing gradient constraints of the physics-based losses.
In contrast to previous work on 3D avatar reconstruction, our method is able to generalize to novel poses with realistic dynamic cloth deformations.
Experiments on several subjects demonstrate that our method can estimate the physical properties of the garments, resulting in superior quantitative and qualitative results compared with previous methods.

\end{abstract}

%%%%%%%%% BODY TEXT
\section{Introduction}
\label{sec:introduction}
Digital human avatars are the backbone of numerous applications in the entertainment industry (e.g., special effects in movies, characters in video games), in e-commerce (virtual try-on), as well as in immersive telecommunication applications in virtual or augmented reality.
Digital human avatars not only have to reassemble the real human in shape, appearance, and motion, but also have to conform to physics.
%
%Hair and 
Clothing must move consistently with the underlying body and its pose.
In the past years, we have seen immense progress in digitizing humans to retrieve such digital human avatars by using neural rendering techniques~\cite{tewari2020neuralrendering,tewari2022advances} or other 3D representations.
Especially, recent methods~\cite{scanimate, POP, DSFN, humannerf, FITE, NPMS, animatablenerf} that rely on deep neural networks to represent appearance and geometry information show promising results. %\TODO{maybe extend this a bit: SZQ: extend "which aims to learn neural mappings between the human pose space and dynamic human avatar representation"?}
%In recent years, methods of digital human reconstruction based on deep learning have increasingly focused on learning realistic dynamic digital human information from real captured data, especially the dynamic deformation of clothing~\cite{scanimate, POP, DSFN}.
%
These data-driven methods store the body and clothing in a unified representation and can be animated using the underlying body prior that exhibits a kinematic deformation tree.
%Previous digital human avatar reconstruction methods have been limited to unified modeling of the human body and clothing, and dynamic simulation of reconstructed digital humans is achieved through skeleton-based animation~\cite{AnimatableHuman} or monocular tracking of body deformations~\cite{DeepCap,livecap}.
%
While the methods aim at generalized animation of the captured human, the results often lack realistic deformations of the garment as they reproduce the deformation states seen during training.
This is due to limited training data, as not all possible poses can be captured as well as occlusions during the scanning procedure.
%
%
%These data-driven methods aim to reconstruct digital humans using scanned data and then generalize them to human poses beyond the dataset.
%However, these methods often suffer from two shortcomings: (1) when there is insufficient scanned data, it can be difficult for these data-driven methods to generalize to untrained human poses; (2) they usually do not incorporate physical constraints or consider the physical characteristics of clothing, resulting in less realistic dynamic clothing models.
%
In contrast to data-driven methods, recent works~\cite{PBNS, SNUG} have focused on incorporating physical constraints into dynamic clothing simulation and generating realistic simulation results of clothing under various complex human poses using unsupervised training.
However, most of these methods rely on a fixed virtual physical model for formulating dynamic clothing, which prevents them from representing the physical properties of real-world clothing from real captures.
%However, most of these methods rely on fixed physical parameters of clothing without the supervision of scanned data, and without the use of real-world clothing information from real captures.
%while the data-driven methods need accurate construction of the clothing physical model to generate a more realistic digital human model.
%

%
In this paper, we propose CaPhy, a clothing simulation and digital human avatar reconstruction method based on an optimizable cloth physical model.
Our goal is to learn realistic garment simulations that can be effectively generalized to untrained human poses using a limited set of 3D human scans.
First, unlike most existing digital human construction works~\cite{scanimate, POP, DSFN}, we model the human body and clothing separately to retain their different dynamic physical properties.
%The proposed dimension-lifting method extracts 3D semantic segmentation of human scans from 2D human parsing results.
%Based on our proposed dimension-lifting method for human 3D semantic segmentation, we solve the naked human body shape and poses from the scans, and obtain the garment templates, respectively.
%
We train a neural garment deformation model conditioned on the human pose that can produce realistic dynamic garment animation results by combining supervised 3D losses and unsupervised physical losses built upon existing real-world fabric measurements.
% To model the dynamic clothing animation, we combine unsupervised training with the losses of a cloth physical model and 3D supervised training of scanned data.
This allows our network to generate simulated clothing results in various human poses even when insufficient scan data is available.
% And it can strengthen the physical constraints of our digital human model when fitting scanned data.
In contrast to SNUG~\cite{SNUG}, we do not assume fixed physical garment properties and optimize the fabric's physical parameters from human scans, thus, capturing its physical characteristics.
By combining the optimization of clothing physical parameters and dynamic clothing training with both physical and 3D constraints,
%unsupervised physical loss and 3D supervised constraints,
our method can generate highly realistic human body and clothing modeling results.
%

% \medskip \noindent
% The contributions of this work are summarized as follows:
% %%%%%%%%%%%%%%%%%%%%%%%%%%%%%%%%%%%%%%%%%%%%%%%%%%%%%%%
% %% contribution 1 and 2: maybe should only keep one? %%
% %%%%%%%%%%%%%%%%%%%%%%%%%%%%%%%%%%%%%%%%%%%%%%%%%%%%%%%
% \begin{itemize}
%     %\item We propose a method for automatically obtaining naked human shape and garment template models from human 3D scans, using our proposed dimension-lifting method for 3D semantic segmentation. 
    
%     \item We reconstruct dynamic clothing models from real-world observations that conform to physical constraints of the input by combining unsupervised training with physics-based losses and supervised training.

%     \item We propose to use real-world fabric measurement results to optimize the state-of-the-art unsupervised physical model for garment simulation.
%     \TODO{are we the first that use this physical model? SZQ: We are first to use that model to construct unsupervised loss}
    
%     \item We optimize the parameters of the cloth physical model to learn the physical properties of clothing from the 3D human scans, which improves the realism of digital human construction.
%     \TODO{that is not really a contribution but necessary for (1)}
% \end{itemize}

\medskip
\noindent
To summarize, we reconstruct dynamic clothing models from real-world observations that conform to the physical constraints of the input by combining unsupervised training with physics-based losses and supervised training. The contributions of this work are as follows:

\begin{itemize}
    \item For garment reconstruction, we introduce a physical model formulation based on real-world fabric measurement results, to better represent the physical properties of real-captured garments (see \cref{sec:network}). 
    \item Using this physics prior and supervised 3D losses, we reconstruct an animatable avatar composed of body and garment layers including a neural garment deformation model which allows us to generalize to unseen poses (see \cref{sec:training}).
    \item Instead of using the given fixed physical parameters of the fabric, we propose to optimize the parameters of the prior to obtain better physical properties of the 3D scans (see \cref{sec:optimization}).
\end{itemize}

\begin{figure*}[t]
    \centering
    \includegraphics[width=\linewidth]{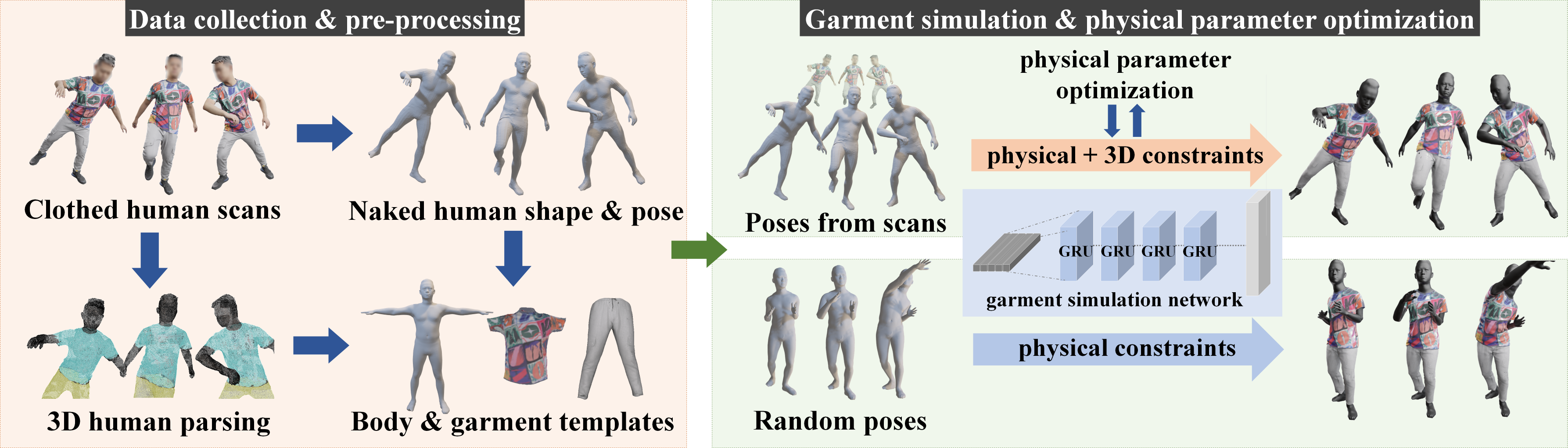}
    \caption{
        \textbf{Overview of CaPhy.} 
        Given a small set of 3D scans of a subject in different poses, the naked body is reconstructed in terms of shape and pose. Static garment templates for the shirt and pants are reconstructed from the scans which are used in the optimization of the neural garment deformation model. Specifically, we optimize for parameters of our physics-based fabric model based on the 3D scans and train a garment simulation network that predicts the deformations conditioned on the pose. Note that the model is also trained on poses different from the scans using the physical constraints of the fabric model.
    }
    \label{fig:method_pipeline}
\end{figure*}

\section{Related Work}
\paragraph{Animatable Human Avatar.} 
Animatable human avatar reconstruction aims to generate a pose-dependent human model based on observations from 3D scans or videos of a particular human object. 
%\TODO{SZQ: keep this? Traditional human avatar methods reconstruct a static digital human model from videos~\cite{VideoAvater,octopus} or a single image~\cite{tex2shape,PixelBased,CAPE,arch++,PIFU,pifuhd,ICON,pamir,photoreal} and can only provide dynamic human geometry by skinning. In recent years, more attention has been paid to learning the dynamic 3D characteristics of the human body and clothing. }
%, especially clothing, from scanned human data in various poses or video inputs, which achieve more realistic 3D human models under dynamic animation. 
Some works focus on building dynamic geometry of human models from scanned data~\cite{scale,scanimate,POP,pointbasedmodel,FITE,neuralgif,UNIF}. Ma et al.~\cite{scale} use local patches to represent the 3D human models, which encode local deformations of different clothing and body parts. 
Saito et al.~\cite{scanimate} employ a reverse skinning network to learn the mapping of the posed space to the canonical space, enabling them to learn the dynamic geometry of clothing in various poses. 
%Li et al.~\cite{avatarcap} propose a human avatar with the subject from a small number of 3D scans, which can be driven by given monocular inputs of the same subject. 
Some methods reconstruct human avatars from a small number of 3D scans~\cite{avatarcap} or a monocular self-rotating human video~\cite{selfrecon}, which can be driven by given monocular RGB inputs of the same subject.
To achieve a more flexible representation and avatar learning, some works utilize point clouds to represent digital humans~\cite{POP,pointbasedmodel,FITE}, while some works leverage SDF (signed distance function) or part-based SDF to represent human avatars~\cite{neuralgif,UNIF}.
%Tiwari et al.~\cite{neuralgif} implicitly reconstruct the shape of clothing in various poses by mapping 3D points in the posed space to the canonical space and predicting the 3D signed distance field (SDF) in the canonical space. 
%Qian et al.~\cite{UNIF} adopt a part-based method that employs implicit fields to learn the unique characteristics of each part of the human body. 
Some works learn the body surface deformation field in the posed space by utilizing single-view RGBD images or depth point clouds, enabling them to describe the dynamic geometry of the body surface~\cite{DSFN,NPMS}. 
These data-driven methods aim to learn the mapping from human pose space to dynamic clothed human models using 3D data. 
However, these methods rely on large-scale 3D training data, posing a challenge for generalization to poses beyond the dataset if insufficient data is available.
%However, these methods rely on large-scale 3D training data, making it challenging to generalize to poses beyond the dataset if insufficient data is available.

More recently, with implicit shape rendering and volume rendering techniques~\cite{nerf,differentiablevolumerendering}, some works focus on generating animatable human avatars with neural rendering methods, 
which supports dynamic human avatar reconstruction from RGB inputs. 
Some methods use neural texture or neural voxel representations~\cite{ANR,neuralbody} to perform neural rendering-based avatars from single- or multi-view images.
%, generating detailed digital human texture rendering results from the rough geometric model of the human body. 
Some methods leverage Nerf representation~\cite{neuralactor,animatablenerf,humannerf,Anerf,hnerf,avatarrex,posevocab} for generating articulated human Nerf models, which propose pose-deformable neural radiance fields for representing human dynamics. 
Grigorev et al.~\cite{stylepeople} integrate neural texture synthesis, mesh rendering, and neural rendering into the joint generation process through adversarial network training. 
Feng et al.~\cite{SCARF} use the mesh-based human model and Nerf-based garment model to better represent different dynamic properties of the human avatar. 
%\TODO{SZQ:intro from SCARF?}
These methods mainly focus on the rendering part without optimizing the dynamic geometry of the clothing.
Therefore, this type of method has limited abilities in capturing detailed wrinkles and physical properties of clothing.

\paragraph{Clothing Capture.} Clothing capture aims to capture the geometric properties of the separate clothing layers. Pons-Moll et al.~\cite{ClothCap} and Tiwari et al.~\cite{SIZER} fit the garment templates to the 3D human scans from the dataset. Jiang et al.~\cite{bcnet} and Chen et al.~\cite{tightcap} propose a method for capturing the body and garment model separately from a single image. Zhu et al.~\cite{deepfashion3d} generate a dataset containing 3D garment models of different styles and learn to reconstruct the 3D shape from 2D images of a garment. Su et al.~\cite{MulayCap} propose a multi-layer human and garment reconstruction method from a monocular RGB video, which recovers the basic garment 3D shapes and captures dynamic garment wrinkle details from the RGB inputs. By learning the 3D semantic field from the pixel-aligned implicit fields, Zhu et al.~\cite{REEF} extract the detailed garment geometry from a single image containing a clothed human. With a single RGBD camera, Yu et al.~\cite{doublefusion} reconstruct the dynamic clothed human model in real-time. Based on this technique, Yu et al.~\cite{SimulCap} separately capture the dynamic human and garment models by combining physical simulations into the garment tracking pipeline. Xiang et al.~\cite{animatablehumanavatarfacebook} utilize a multi-view capture system consisting of about 140 cameras to collect high-resolution videos for reconstructing the human body in clothing. 
In addition, some researchers focus on establishing correspondence between deformed garments~\cite{garmentalignment}.
Some researchers focus on more flexible garment shape and style representation and capture using UV-based methods~\cite{deepcloth_su2022,tex2shape,PixelBased} or deep
unsigned distance functions~\cite{SIMPLicit}. 
These clothing capture methods focus on capturing the static or dynamic geometric features of clothing while disregarding the extraction of physical properties of garments and the application of physically realistic garment animation with the captured garments.

\paragraph{Physical-based Garment Simulation.}
Physics-based simulation methods are widely used in dynamic garment reconstruction.
Traditional physics-based garment simulation methods~\cite{Provot95deformationconstraints,bonet1997nonlinearcontinuummechanics,Anisotropic} rely on force and collision modeling or explicit time integration. 
Thus, applying these methods to clothing geometry inference using neural networks or integrating them with data-driven methods can be challenging, making them difficult to incorporate into a human avatar generation framework. 
Typical neural network-based methods for garment simulation~\cite{TailorNet,virtualtryon,virtualtryongnn,virtualbonecloth}
%, such as Patel et al.~\cite{TailorNet}, Santesteban et al.~\cite{virtualtryon} and Vidaurre et al.~\cite{virtualtryongnn}, 
rely on pre-generated virtual clothing simulation data using pre-defined simulators, without incorporating physical models into the deep learning framework.
Recent research has made breakthroughs in developing physical constraints for clothing using neural networks.
Bertiche et al.~\cite{PBNS} incorporate physical constraints into the loss function by imposing both edge distance constraints and bending constraints, which achieves the first unsupervised training of a physical simulation model for clothing using neural networks.
Santesteban et al.~\cite{SNUG} employ the StVK (Saint Venant-Kirchhoff) model to build the neural physical model of clothing, which further improves the realism of the animated garments. 
However, these methods rely on fixed physical parameters of the fabric during training as well as require a fixed template mesh, with limited ability to represent the geometrical and physical properties of real captured garments.

\section{Method}
\label{sec:method}
%In this paper, we propose a %garment simulation and 
%realistic human avatar reconstruction pipeline based on static 3D human scans. 
Our goal is to extract the geometrical and physical characteristics of specific garments
from a limited set of 3D scans, typically consisting of 50 to 120 scans of a clothed human in various poses.
We propose to learn a neural deformation model which imitates a garment simulation, to extrapolate to novel poses of the digital human avatar.
%We first introduce a dimension-lifting method for extracting 3D semantic segmentation from 2D human parsing results, which enables labeling body and garment vertices from the 3D scans. 
%, which enables accurate 3D semantic segmentation of corresponding human body scans.
In the first step, we extract the naked human and garment templates from the static scans, leveraging 2D semantic segmentations (see \cref{sec:reconstruction}).
%Based on the 3D semantic segmentation, we solve the naked human body shape and poses, and obtain the garment templates, respectively (see \cref{sec:reconstruction}).
%Specifically, we use the deformed SMPL-X~\cite{SMPL-X} model to represent the real human body underneath the clothing. The garment templates are obtained by sampling the garment style parameter from the style space in~\cite{TailorNet}.
%Furthermore, we propose unsupervised physical losses for animating 3D garments by using a cloth physical model, which is built upon existing real-world fabric measurements.
%Combined with the 3D matching loss using 3D scanned data, 
Furthermore, by combining supervised 3D losses and unsupervised physical losses built upon existing real-world fabric measurements, we train a garment simulation model that can produce realistic dynamic garment animation results, which also shows 3D consistency with the scans (see \cref{sec:network,sec:training}).
Based on such a model, we propose a method for optimizing the physical parameters of the cloth prior, such that the animated garment has consistent physical characteristics with the scanned data (see \cref{sec:optimization}).
%Finally, a realistic digital human avatar based on static human scans is established.
To alleviate the collision between the upper and lower body garments, we fine-tune the model constrained by a collision loss (see \cref{sec:collision}).

%%%%%%%%%%%%%%%%%%%%%%%%%%%%%%%%%%%%%%%%%%%%%%%%%
%\subsection{Human model and Garment Template Reconstruction}
\subsection{Data Pre-processing}
\label{sec:reconstruction}
Given 3D scans of a clothed human in various poses, we extract the naked human model and the basic garment templates.
Specifically, we use the SMPL-X~\cite{SMPL-X} model to represent our human body.
SMPL-X is a parametric model that represents the body, face, and hands jointly with the pose ($\theta$) and shape ($\beta$) parameters.
By applying deformations to the base SMPL-X model, we can represent human faces and other surface details:
\begin{equation}
\label{eq:smplx}
    \mathcal{T}(\beta,\theta,dT)=W(T(\beta)+dT,J(\beta),\theta,w)
\end{equation}
where $W(\cdot)$ is a standard linear blend skinning function, $T(\beta)$ is the base SMPL-X template mesh in T-pose parameterized by $\beta$, and $T(\beta)+dT$ adds corrective vertex displacements $dT$ on the template mesh.
$J(\beta)$ outputs 3D joint locations from the human mesh.
$w$ are the blend weights of the skeleton $J(\beta)$.

\begin{figure}[t]
    \centering
    \begin{subfigure}[b]{0.115\textwidth}\includegraphics[width=\textwidth]{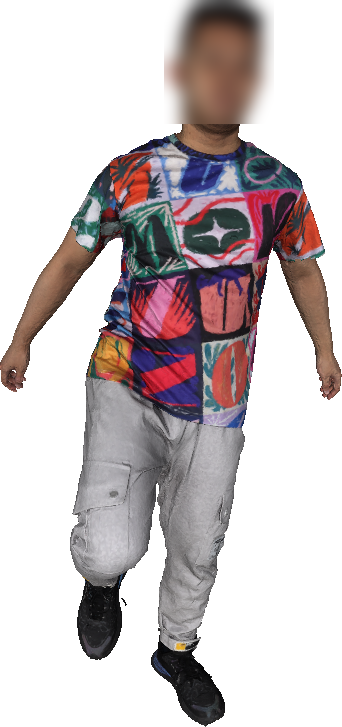}\caption{}\end{subfigure}
    \begin{subfigure}[b]{0.115\textwidth}\includegraphics[width=\textwidth]{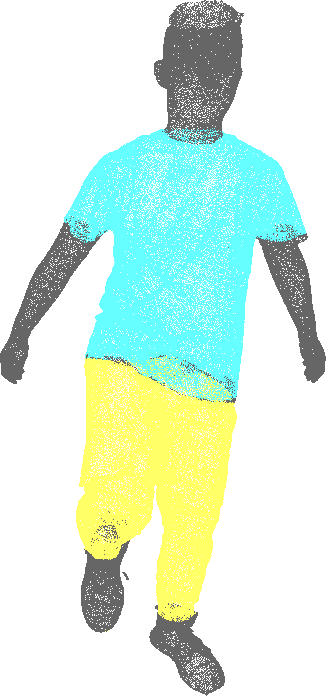}\caption{}\end{subfigure}
    \begin{subfigure}[b]{0.115\textwidth}\includegraphics[width=\textwidth]{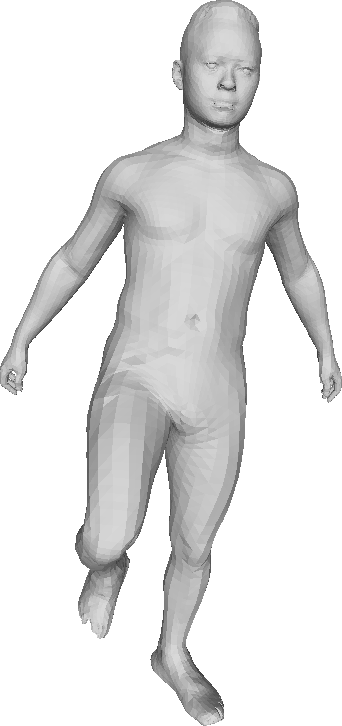}\caption{}\end{subfigure}
    \begin{subfigure}[b]{0.115\textwidth}\includegraphics[width=\textwidth]{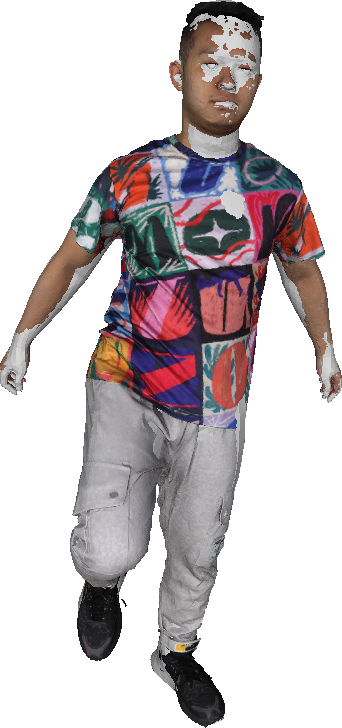}\caption{}\end{subfigure}
    \caption{(a) The scanned human data. (b) 3D semantic segmentation of the scan. (c) The generated naked human body shape. (d) The results of fitting (c) to (a).}
    \label{fig:solvebody}
\end{figure}

\paragraph{Human Body Shapes and Poses.}
To obtain the naked human shape $T(\beta)+dT$ and pose $\theta_i^V$ of the human scan $\mathcal{V}_i$, 
%a common solution would be to use the iterative closest point (ICP) algorithm for solving the base shape parameters $\beta$ and pose parameters $\theta_i^V$, and to use non-rigid ICP for solving the per-vertex deformation $dT$ in the canonical space. However, these methods are not suitable for our case, as our scanned data does not consist of minimally clothed individuals. Therefore, 
we leverage 2D semantic information of the scans to solve the ``naked body underneath the clothing'' problem.
Specifically, for a 3D human scan $\mathcal{V}$, we first render it with 32 uniformly distributed viewing angles to obtain color images $C^{(j)}$ and depth images $D^{(j)}(j=1,2,...,32)$.
The color images are used for obtaining 2D semantic segmentation results, while the depth images are used for determining the visibility of each scan vertex at each viewing angle.
We apply \cite{SelfCorrHumanParsing} to color images $C^{(j)}$ to obtain 2D semantic segmentation values. %$\mathcal{H},\mathcal{G}_u,\mathcal{G}_l$, where $\mathcal{H}$ represents the body surface, and $\mathcal{G}_u$ and $\mathcal{G}_l$ represent upper and lower garments, respectively.
%To simplify notation, we use $\mathcal{G}$ to represent the garments below.
For each vertex $\mathbf{v}\in \mathcal{V}$, we project it to each view $j$ to get its corresponding projected depth. We consider a vertex to be visible in view $j$ if its projected depth approximates the corresponding depth value in the depth image $D^{(j)}$.
%For each visible vertex in a given view $j$, we extract the corresponding 2D semantic segmentation value from the color image $C^{(j)}$.
We then assign the 3D semantic segmentation value of the vertex $seg_{\mathbf{v}}\in\{\mathcal{H},\mathcal{G}_u,\mathcal{G}_l\}$,
by applying a majority rule based on the 2D semantic segmentation values extracted from all visible viewing angles, where $\mathcal{H}$, $\mathcal{G}_u$ and $\mathcal{G}_l$ denote body and different garment labels, respectively. As shown in \cref{fig:solvebody}(b), we obtain accurate 3D human semantic segmentation results for the scan.

To incorporate this 3D semantic segmentation information for solving the naked human body underneath clothing, for each deformed SMPL-X vertex $t\in{\mathcal{T}(\beta,\theta,dT)}$, we modify the vertex fitting energy term in both the ICP (iterative closest point) solver for calculating shape $\beta$ and pose $\theta_i^V$ and in the non-rigid ICP solver for vertex displacements $dT$, by constraining the deformed vertices to align the uncovered skin areas while being underneath the clothing areas.
%as follows:
% \begin{equation}
% \label{eq:icp_fitting_term}
%     E_\mathbf{t}=\left\{
%     \begin{aligned}
%         ||\mathbf{n}_\mathbf{t} \cdot (\mathbf{t}-\mathbf{v}_\mathbf{t})||^2 \quad if \quad seg_{\mathbf{v}_\mathbf{t}}=\mathcal{H}\\
%         ||max(0, \mathbf{n}_\mathbf{t} \cdot (\mathbf{t}-\mathbf{v}_\mathbf{t}+\delta))||^2 \quad if \quad seg_{\mathbf{v}_\mathbf{t}} = \mathcal{G}\\
%         0 \quad \quad \quad  otherwise
%     \end{aligned}
%     \right.
% \end{equation}
% where $\mathbf{v}_\mathbf{t}$ represents the nearest human scan vertex used to constrain vertex $\mathbf{t}$, while $\mathbf{n}_\mathbf{t}$ denotes its normal. 
% We set $\delta=5mm$ when the label of $\mathbf{v}_\mathbf{t}$ denotes garments.
As shown in \cref{fig:solvebody}(c)(d), the mesh of the body closely conforms to the original scan.

\paragraph{Static Garment Templates.}
Our goal is to generate static garment templates consistent with the scans containing less dynamic garment geometry information such as folds, which can be used in our physics-based training.
%To obtain a garment template from the input human scans to be used in our 
%, our goal is to generate static garment templates containing less dynamic garment geometry information such as folds.
%
To this end, we use TailorNet~\cite{TailorNet} for generating our garment template model in a standard pose.
%We use the garment shape parameter $\gamma$ to control the shape of a specific garment, such as the length of the sleeves or the clothing width of the torso.
%
For the upper and lower garment (e.g., shirts and pants), we use the same procedure.
In the following, we will describe it for the upper garment generation.
We first select one scan $\mathcal{V}^S$ that is closest to the standard T-pose or A-pose from our scanned data, and then extract the upper body garment vertices $\mathcal{V}_{\mathcal{G}}^S=\{\mathbf{v}|\mathbf{v}\in \mathcal{V}^S,seg_\mathbf{v}=\mathcal{G}_u\}$ by leveraging the fused 3D semantics.
We sample 300 garment shape parameters $\gamma_i(i=1,2,...,300)$ from the garment style space in~\cite{TailorNet} to generate 300 upper body garments $\mathcal{V}_{\mathcal{G}}^i$.%(i=1,2,...,300)$.
After calculating the 3D Chamfer distance between $\mathcal{V}^S$ and $\mathcal{V}_{\mathcal{G}}^i$ for all 300 samples, we select the generated garment with the smallest Chamfer distance and set it to the static garment template $\mathcal{V}_{\mathcal{G}}^T$.
\subsection{Garment Simulation Network}
%\TODO{try to have a consistent naming; either garment simulation network, or neural garment deformation model}
\label{sec:network}
After obtaining the human and garment templates, we train a garment simulation network from the scans.
Our deformation model for the clothing is defined as:
\begin{equation}
\label{eq:skincloth}
    \mathcal{V}_{\mathcal{G}}^{anim}(\theta)=W(\mathcal{V}_{\mathcal{G}}^T+d\mathcal{V}_{\mathcal{G}}(\theta),J(\beta),\theta,w_{\mathcal{G}}) ,
\end{equation}
where $w_{\mathcal{G}}$ is the skinning weights of the static garment template $\mathcal{V}_{\mathcal{G}}^T$. 
$w_{\mathcal{G}}$ on each garment vertex is determined by the nearest-neighbor body vertex.
Other symbols remain the same as defined in \cref{eq:smplx}.
$d\mathcal{V}_{\mathcal{G}}(\theta)$ represents the garment deformation in the canonical space.
We aim to learn the garment deformation $d\mathcal{V}_{\mathcal{G}}(\theta)$ as a function of the human body poses $\theta$.
The garment deformation is predicted in the canonical space as in SCANimate~\cite{scanimate} and SNUG~\cite{SNUG}.

%To construct the dynamic garment network, we begin with the training network described in~\cite{SNUG}, which is based on unsupervised physical losses.
We construct the garment simulation network as in SNUG~\cite{SNUG}, which proposes an unsupervised garment dynamic learning pipeline.
Specifically, the garment simulation network structure consists of a 4-layer GRU network and a fully connected layer, which inputs 5 continuous human poses $\theta_i^{(1,2,3,4,5)}$ and outputs the dynamic garment deformations $d\mathcal{V}_{\mathcal{G}}(\theta_i^{(1,2,3,4,5)})$. 
The model also supports inputs from static poses.
%To enable the network to learn from static poses, we replicate the independent human pose $\theta_s$ 5 times as inputs. \TODO{unclear}

Similar to Santesteban et al.~\cite{SNUG}, we use unsupervised physical losses, but additionally we add 3D matching losses to the network training to match the physics of the real observations.
Santesteban et al.~\cite{SNUG} propose the physics-based losses using the Saint Venant Kirchhoff (StVK) model, where the strain energy is formulated as:
%The Saint Venant Kirchhoff (StVK) model defines membrane strain energy in~\cite{SNUG}, which constructs the unsupervised physical losses.
%The StVK model is formulated as:
\begin{equation}
\label{eq:stvk}
    E_{strain}=V(\frac{\lambda}{2}tr(G)^2+\mu tr(G^2)) ,
\end{equation}
where $V$ is the volume of each triangle.
$\lambda$ and $\mu$ are the Lam\'{e} constants, and $G$ is the Green strain tensor:
\begin{equation}
    G=\frac{1}{2}(F^TF-I),
\end{equation}
where $F$ is the local deformation metric of a triangle.
However, we find that the StVK model with constant $\lambda$ and $\mu$ does not accurately capture the strain behavior of real fabrics under tensile deformation.
On the other hand, Wang et al.~\cite{arcsim11DDE} measures $\lambda$ and $\mu$ for 10 different cloth materials and find that they are related to the principal strains and strain angles, thus related to Green strain tensor $G$.
Therefore, we follow~\cite{arcsim11DDE} and rewrite \cref{eq:stvk} to establish a real-measured anisotropic strain model as follows:
\begin{equation}
\label{eq:arcsim}
    E_{strain}=V(\frac{\lambda^m(G)}{2}tr(G)^2+\mu^m(G) tr(G^2)) ,
\end{equation}
where $\lambda^m(G)$ and $\mu^m(G)$ are obtained by interpolating the results measured by~\cite{arcsim11DDE} under several $G$ conditions.
$m=1,2,...,10$ represent 10 different cloth materials.
We then define our physics-based losses similar to~\cite{SNUG} as:
\begin{equation}
\label{eq:phys}
\begin{split}
    E_{phys}=\sum_F E_{strain}+\sum_e E_{bend}+\sum_{\mathbf{v}} E_{gravity} \\ 
    +\lambda_{collision}\sum_{\mathbf{v}}E_{collision}+\sum_{\mathbf{v}}E_{dyn} ,
\end{split}
\end{equation}
where %$E_{strain}$ represents the membrane strain energy of garments, as mentioned above, and 
$E_{bend}$ models the bending energy determined by the angle of two adjacent triangles.
Here we also adjust the bending term according to ARCSim~\cite{arcsim11DDE,arcsim14} as follows:
\begin{equation}
    E_{bend}=\frac{l^2k^m_{bending}}{8A}(\tau-\tau_T)^2 ,
\end{equation}
where $k^m_{bending}$ is the bending stiffness coefficient of the cloth material $m$~\cite{arcsim11DDE}, $\tau$ and $\tau_T$ denote the angles of two adjacent triangles under animation and static, respectively. $l$ and $A$ denote the adjacent edge length and the sum of the adjacent triangle areas. Other terms in \cref{eq:phys} are formulated similarly to~\cite{SNUG}.
% $E_{gravity}$ represents the gravity energy of each clothing vertex, $E_{collision}$ is the collision constraint between clothing and body.
% We formulate $E_{gravity}$ and $E_{collision}$ as follows:
% \begin{equation}
% \label{eq:gravity_collision}
% \begin{aligned}
%     E_{gravity}&=m_{\mathbf{v}}gy_{\mathbf{v}} \\
%     E_{collision}&=||\max (\mathbf{n}_b \cdot(\mathbf{v}-\mathbf{v}_b - \delta), 0)||^3
% \end{aligned}
% \end{equation}
% where $m_{\mathbf{v}}$ and $y_{\mathbf{v}}$ are the average quality and height of each clothing vertex $\mathbf{v}$, $\mathbf{v}_b$ and $\mathbf{n}_b$ are the nearest neighbor body vertex and corresponding normal.
% $\delta=5mm$ represents the minimum distance between the clothing and body vertices.
% $E_{dyn}$ constructs the kinetic energy constraint of continuous motion of clothing at each vertex:
% \begin{equation}
%     E_{dyn}=\sum_{t=1}^{3}\frac{1}{2\Delta t}m_{\mathbf{v}}|||v^{(t+2)}-2v^{(t+1)}+v^{(t)}||^2
% \end{equation}
% where $v^{(t)}$ represents the clothing vertex coordinates of the $t$ frame, $\Delta t$ indicates the time interval between two frames.

\subsection{Training with physics and 3D constraints}
\label{sec:training}
In order to make our simulated garments both have physical realism and conform to our 3D scans, we introduce the combined training for garment simulation with both unsupervised physical loss and supervised 3D loss, which establishes the connection between physical-based simulated virtual garments and garments from real-world captures.

For the training using the physics-based loss, we randomly sample human poses from the CMU Mocap dataset.
Specifically, we use 10000 sets of 5 consecutive frames of human poses $\theta_i^{(1,2,3,4,5)}(i=1,2,...,10000)$.
Here, we select the appropriate fabric material parameters according to the material descriptions in~\cite{arcsim11DDE} and the type of clothing in the data.
To ensure that our model can also conform to our 3D scanned data, we also train the garment simulation network  for the human poses $\theta^V$ of the 3D scans $\mathcal{V}$ generated in \cref{sec:reconstruction} by minimizing the following loss function:
\begin{equation}
\label{eq:supervised}
    E_{garment}=E_{phys}+\lambda_{3d}E_{3d} ,
\end{equation}
where $E_{phys}$ is the physics-based loss defined in \cref{eq:phys}, and $E_{3d}$ is the supervised 3D matching loss defined as:
\begin{equation}
\begin{aligned}
    E_{3d}=\frac{1}{M}\sum_{\begin{subarray}{c} i=1 \\ v_i\in\mathcal{V}_{\mathcal{G}}^{anim}(\theta^V)\end{subarray}}^M\min_{v_j\in \mathcal{V}_{\mathcal{G}}}||v_i-v_j||^2+ \\
    \frac{1}{N}\sum_{\begin{subarray}{c} j=1 \\ v_j\in\mathcal{V}_{\mathcal{G}}\end{subarray}}^N\min_{v_i\in \mathcal{V}_{\mathcal{G}}^{anim}(\theta^V)}||v_j-v_i||^2 ,
\end{aligned}
\end{equation}
which computes the 3D Chamfer distance between the generated garments $\mathcal{V}_{\mathcal{G}}^{anim}(\theta^V)$ and the ground truth garment point cloud $\mathcal{V}_{\mathcal{G}}$ extracted from scan $\mathcal{V}$.

During training, in order to balance the unsupervised training for random poses and the supervised training for scan poses, we introduce a 1:4 training strategy, where we train our network for 1 epoch with random poses and 4 epochs with scan poses, repeatedly.
In our experiments, we find that such a training strategy leads to natural and physically realistic garment dynamic results for both scan data and randomly sampled poses.

\begin{figure}
    \centering
    \includegraphics[width=\linewidth]{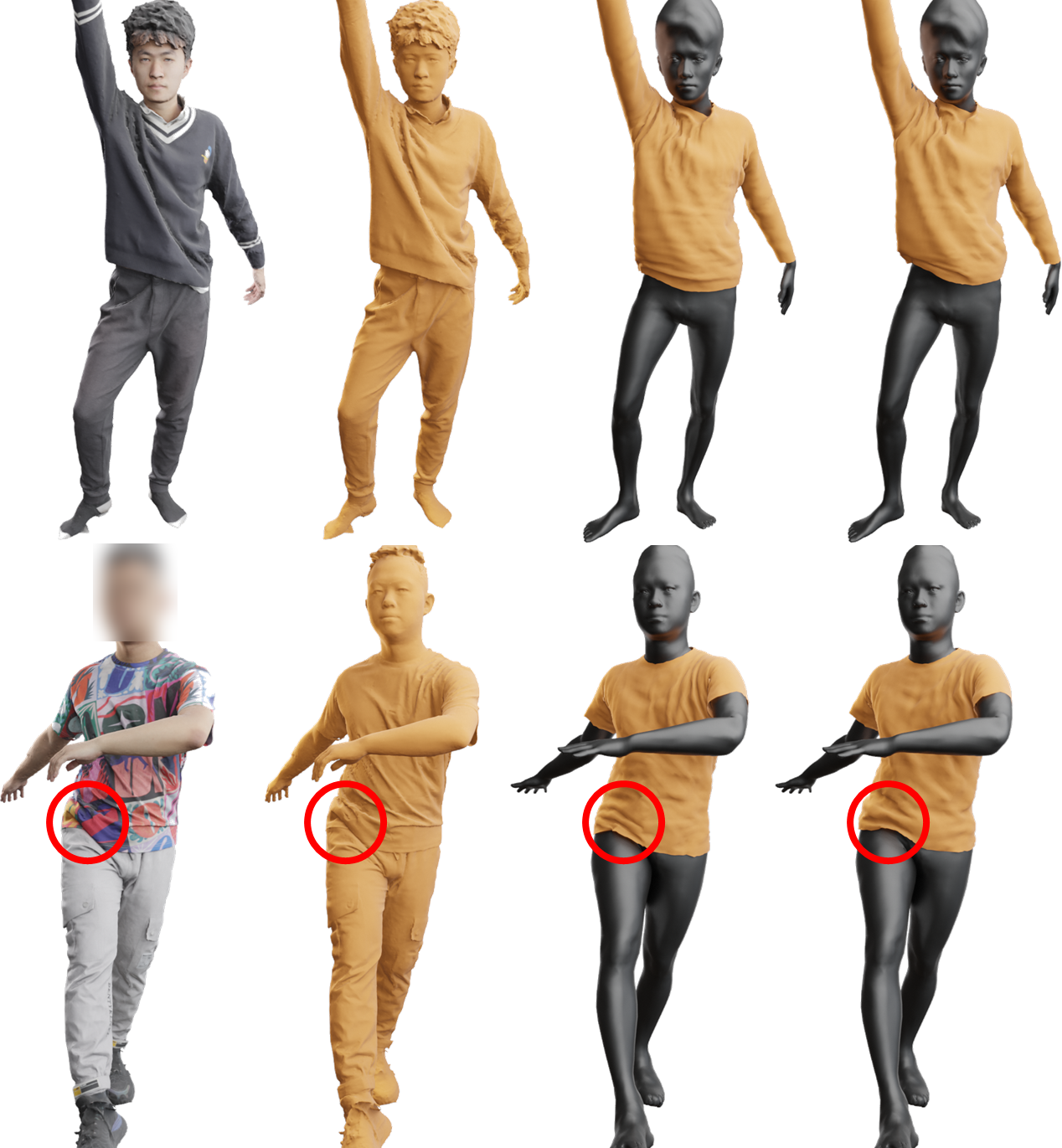}
    % {\includegraphics[width=0.2\linewidth]{fig/optphys1.png}}
    % {\includegraphics[width=0.2\linewidth]{fig/optphys1.png}} % TODO: geometry here
    % {\includegraphics[width=0.2\linewidth]{fig/optphys2.png}}
    % {\includegraphics[width=0.2\linewidth]{fig/optphys3.png}}
    %
    \caption{
    Garment animation results for test set models. 
    From left to right: ground truth color and geometry, our results before and after physical parameter optimization.% of the cloth.
    %The left two models represent the ground truth human scan with and without texture, while the right two models show the animation results before and after physical parameter optimization of the cloth.
    %\TODO{can we show the GT geometry here?}
    }
    \label{fig:optphys}
\end{figure}

\subsection{Cloth Physical Parameter Optimization}
\label{sec:optimization}
As mentioned in \cref{sec:network}, for training a physically realistic garment simulation model, we select appropriate fabric material parameters to build our unsupervised physical losses. However, these preselected fabric parameters may not reflect the actual physical properties of the scan data. %Thus, we propose a method for optimizing the physical parameters of clothing, to further extract the physical characteristics of garments from the scans.

Thus, our goal is to optimize the fabric's physical parameters using the scanned data as ground truth.
%We use the selected cloth physical parameters as the initial state.
We observe that after training, the generated garments $\mathcal{V}_{\mathcal{G}}^{anim}(\theta^V)$ under scan pose $\theta^V$ converge to the global minimum of the sum of physical constraint energy and 3D constraint energy. 
As a result, we hypothesize that the gradients of the physical constraint energy and the 3D constraint energy of $\mathcal{V}_{\mathcal{G}}^{anim}(\theta^V)$ are opposite for each vertex.
And ideally, if the physical model matches the real garments, the gradients of the two energies should approximate zero when the generated garments coincide with scans.
Under such assumptions, we iteratively train the garment simulation network and optimize the fabric's physical parameters. We first fix the generated garments $\mathcal{V}_{\mathcal{G}}^{(0)}(\theta^V)=\mathcal{V}_{\mathcal{G}}^{anim}(\theta^V)$ generated by our pre-trained garment simulation model, and then optimize the physical parameters.
%The generated garments $\mathcal{V}_{\mathcal{G}}^{(0)}(\theta_i^V)$  are only determined by the network outputs $d\mathcal{V}_{\mathcal{G}}^{(0)}(\theta_i^V)$ when the human pose $\theta_i^V$ is fixed.
The optimized physics-based loss term for dynamic garments is denoted as $E_{phys}^{(1)}$.
Then we use \cref{eq:supervised} to train the garment network and obtain $\mathcal{V}_{\mathcal{G}}^{(1)}(\theta^V)$, etc. After the iterative training, the model can generate dynamic garments that are geometrically and physically consistent with the scan data.

In our pipeline, we denote the fabric's physical parameters as $\Theta_G$, which correspond to the measured parameters in~\cite{arcsim11DDE} used for calculating $\lambda^m(G)$ and $\mu^m(G)$, and $k_{bending}^m$, as mentioned in \cref{sec:network}.
The loss function of the physical parameters optimization is defined as follows:
\begin{equation}
    E_{param}=||\frac{\partial E_{phys}^{(j)}(\Theta_G)}{\partial d\mathcal{V}_{\mathcal{G}}^{(j)}(\theta^V)}||^2+\lambda_{reg}^\Theta||\delta \Theta_G||^2 ,
\end{equation}
where $\frac{\partial E_{phys}^{(j)}(\Theta_G)}{\partial d\mathcal{V}_{\mathcal{G}}^{(j)}}$ represents the gradient constraint of the physical energy constraint $E_{phys}^{(j)}$ on the outputs of the garment dynamic network after $j$ iterations.
$\delta \Theta_G$ is the physical parameter update of each iteration, which is constrained by the regularization coefficient $\lambda_{reg}^\Theta$. We adjust $\lambda_{reg}^\Theta$ term for different parameters and reparametrize $k_{bending}^m$ for balancing the order of magnitude between different parameters. 
We perform physical parameter optimization with the scan poses, and then train our garment simulation network with both random poses and scan poses with the optimized physics-based losses.
\cref{fig:optphys} shows that the animation results of the garments more closely resemble the physical wrinkle details of the ground truth after optimization of the fabric's physical parameters. 

\begin{figure}
    \centering
    \includegraphics[width=\linewidth]{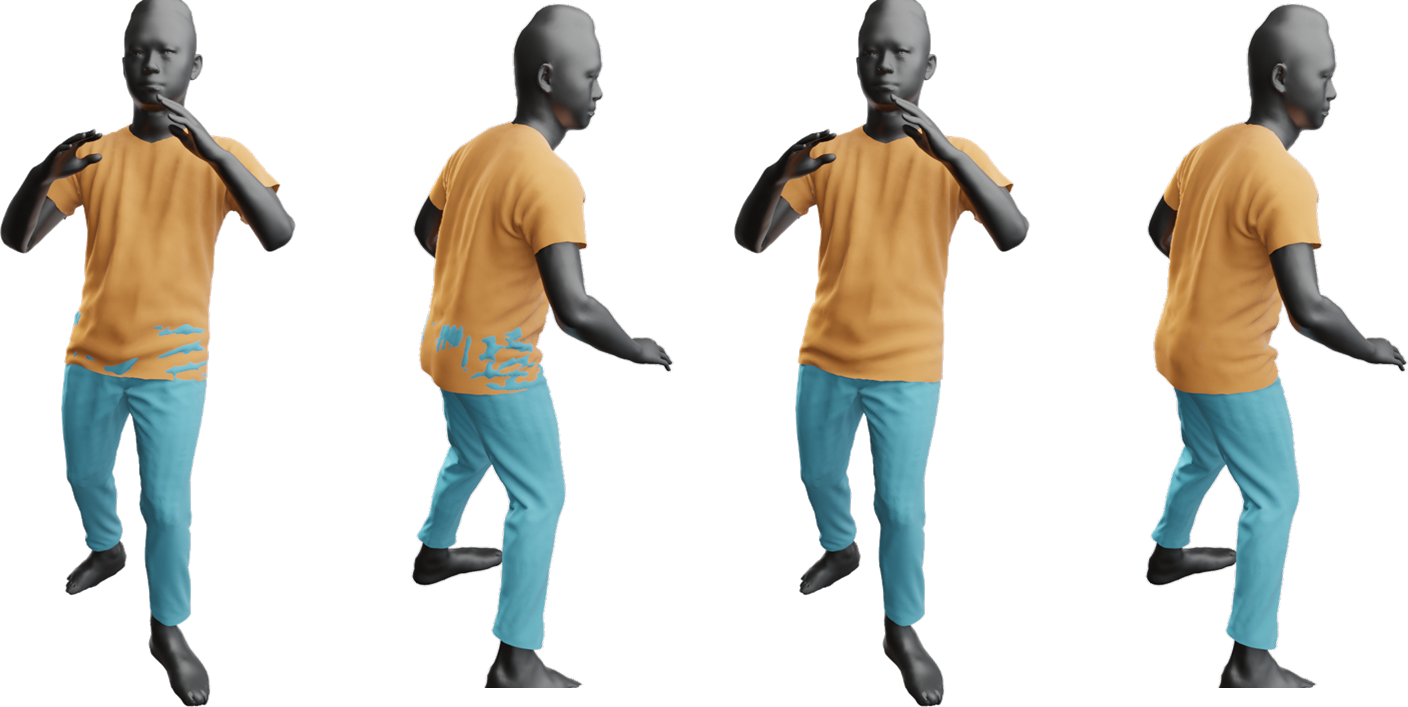}
    \caption{Collision Fine-tuning: The left two are without fine-tuning, while the right two results are with fine-tuning.}
    \label{fig:collision}
\end{figure}

\subsection{Collision Fine-tuning}
\label{sec:collision}

After the physical parameter optimization and garment simulation training step, we can generate vivid animation results for each garment of the scanned person.
However, we have not considered the penetration of different garments. Therefore, we perform a collision fine-tuning training step to address the collision between the upper and lower garments, as shown in \cref{fig:collision}. See more results in the supplementary.

%To address the collision between the upper-body and lower-body garments, we add the collision constraint loss term to fine-tune the trained network.
Specifically, after training for each garment, we fine-tune the dynamic networks of the upper and lower garments using random pose data with the loss function:
\begin{equation}
\begin{split}
    E_{finetune}=E_{phys(upper)}+E_{phys(lower)} \\
    +\lambda_{colli}E_{colli(upper-lower)} .
\end{split}
\end{equation}
For the scanned data, we add $E_{3d(upper)}$ and $E_{3d(lower)}$ accordingly. The collision term is defined similarly to the garment-body collision term in \cref{eq:phys}: 
\begin{equation}
    E_{colli(upper-lower)}=||\max (\mathbf{n}_l \cdot(\mathbf{v}_u-\mathbf{v}_l - \delta), 0)||^3 ,
\end{equation}
where $\mathbf{v}_l$ is the nearest lower garment vertex of each upper garment vertex $\mathbf{v}_u$. $\mathbf{n}_l$ is the corresponding normal.
We only add constraints to the upper garment region near the lower garment, and fine-tune the network alternately on the random pose and scanned pose data for 10 epochs.
%As shown in \cref{fig:collision}, the interpenetration between upper and lower garments is eliminated after the collision fine-tuning training.

\begin{figure*}[t]
    \centering
    \includegraphics[width=\linewidth]{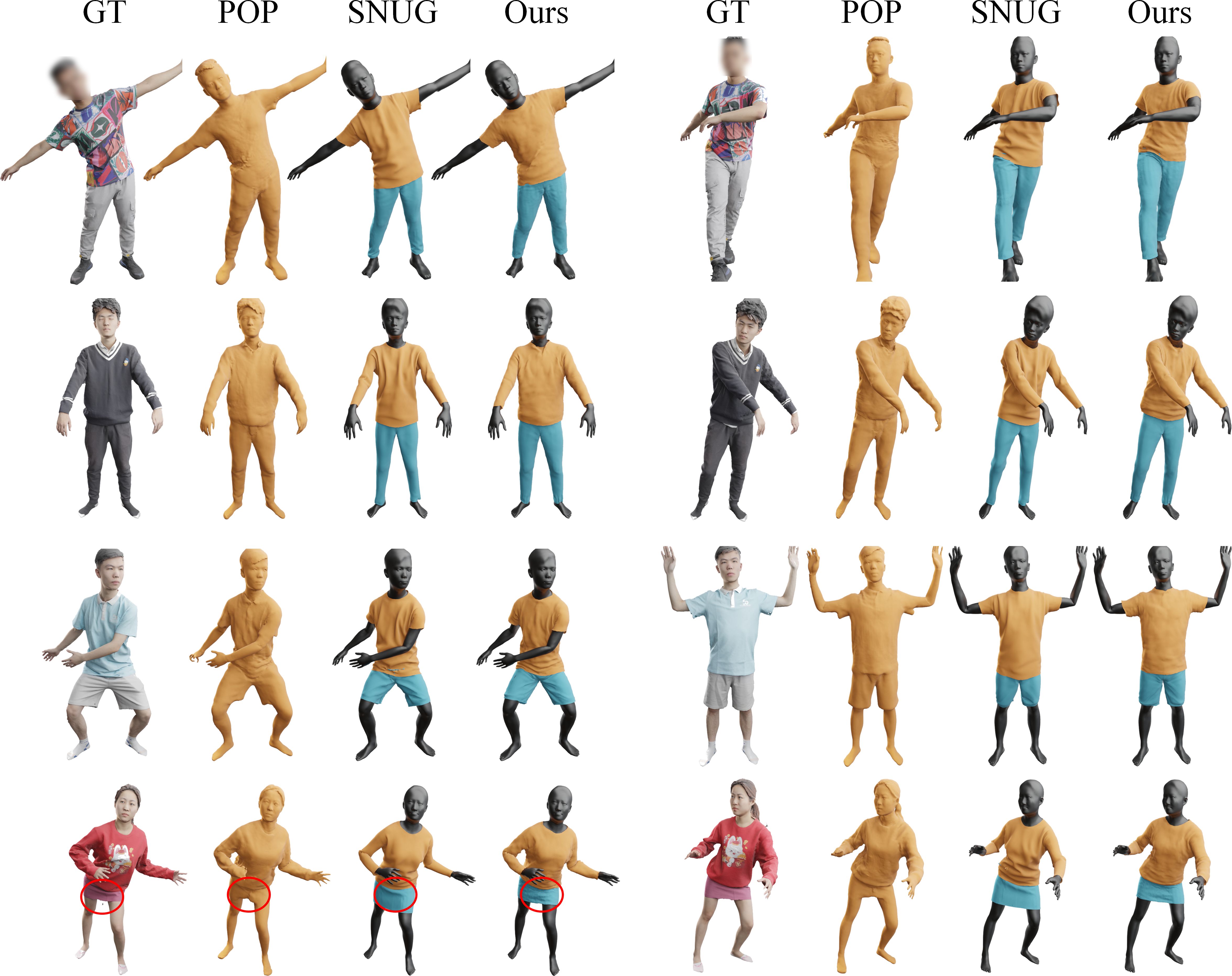}
    \caption{Animation results of the reconstructed digital humans using different methods. Each group from left to right: ground truth 3D scans from the test set, POP~\cite{POP} results, SNUG~\cite{SNUG} results and our results.%The left four columns show the animations on the test set, with poses not existing in the training procedure. while the right three columns display results using random poses. \TODO{add the additional subjects, merge with comparison figure}
    }
    \label{fig:results}
\end{figure*}

\section{Experiments}
\label{sec:experiments}
In our experiments, we employ a multi-camera system to capture 3D scans of clothed human bodies, which is used to evaluate the effectiveness of our proposed method.
For each %set of clothing of each 
collected human subject, we use 80\% of the collected data as the training set and the remaining 20\% as the test set.
%For the test set, we only perform reconstruction of human body shapes and poses in \cref{sec:reconstruction} to test the accuracy of digital human reconstruction in a given pose.
In addition, we show dynamic animations of the reconstructed digital human models based on monocular video tracking.

\textbf{Training:}
In addition to data preparation and template reconstruction for the human and garments, each garment requires approximately 6-11 hours for the optimization of the fabric's physical parameter described in \cref{sec:optimization}, 8-14 hours for dynamic garment network training (\cref{sec:training}), and 1 hour for collision fine-tuning (\cref{sec:collision}), using a single NVIDIA RTX 3090 GPU.
The inference module takes about 3-6 ms per garment per pose (depending on the size of the garment template) which in principle could allow interactive control of the reconstructed human avatar.

\subsection{Dynamic Animation Results}
\label{sec:results}
\begin{figure}[t]
    \centering
    \includegraphics[width=\linewidth]{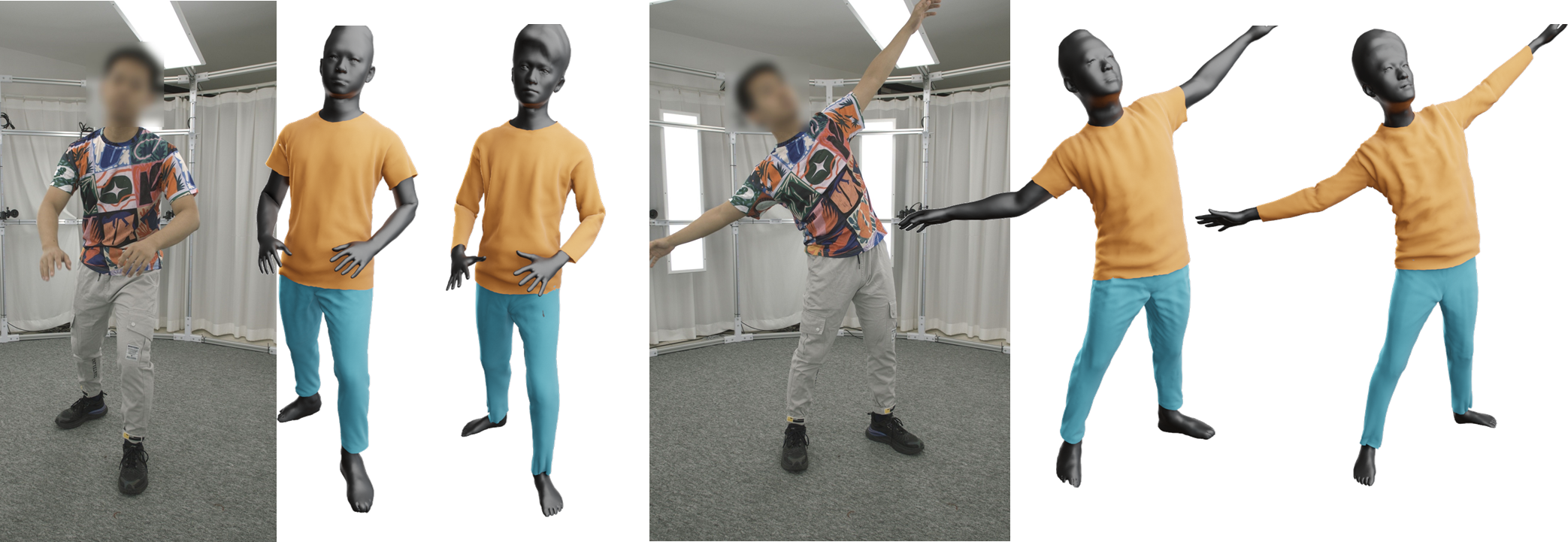}
    \caption{The animation results from single-view videos applied to two reconstructed human avatars.}
    \label{fig:results_video}
\end{figure}

\begin{figure}[t]
    \centering
    \includegraphics[width=\linewidth]{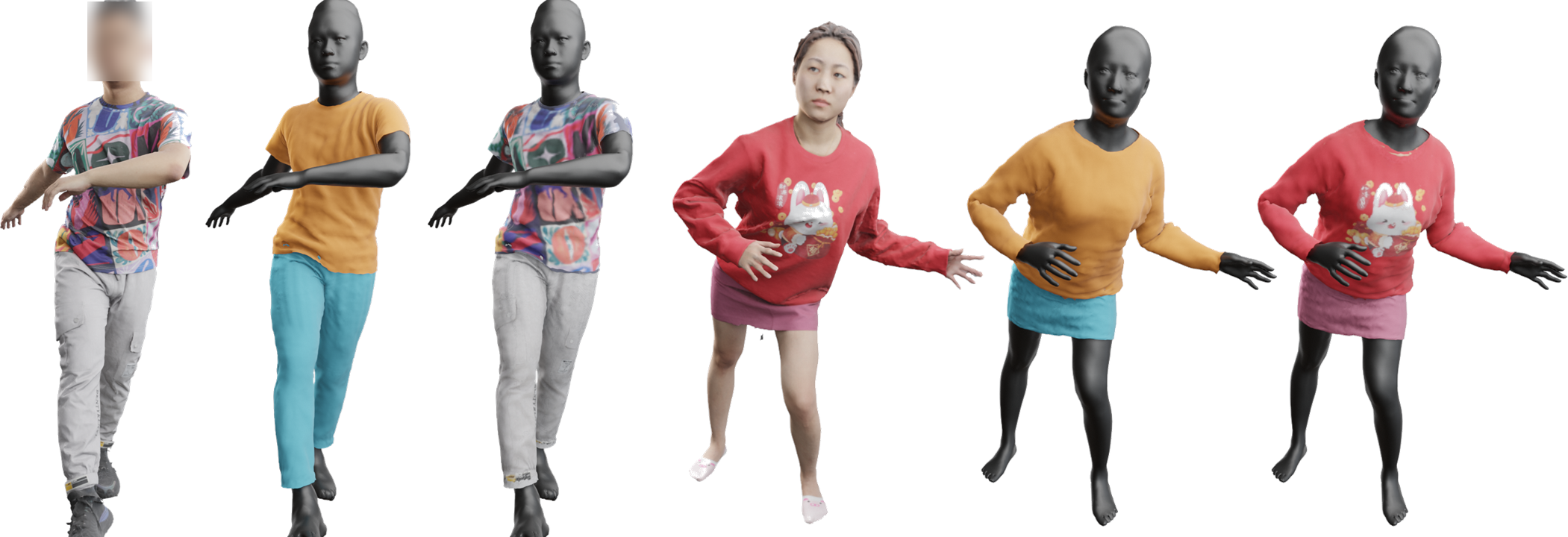}
    \caption{Two examples of texture reconstruction results. Each from left to right: the test set model, results without texture, and results with texture.}
    \label{fig:results_texture}
\end{figure}

To evaluate our method, we reconstruct different digital humans from collected human scans with different styles of clothing. 
As shown in \cref{fig:results}, the generated digital human models accurately replicate the physical characteristics of the scanned human subjects' clothing on the test set, including wrinkles. Note that our results are visually more similar to the ground-truth observations in our test set than previous methods.
\begin{comment}
Moreover, benefiting from the physical constraints in training, the reconstructed digital human models can generate clothing animation results with physical realism for poses in the CMU Mocap dataset.
Note that the clothing animation results show different physical characteristics for different clothing of different human scans.    
\end{comment}
%
The reconstructed digital human models can also be driven by a monocular RGB video with the human poses estimated by PyMAF~\cite{pymaf}.
As shown in \cref{fig:results_video}, we generate digital human animation results based on a single-view video, similar to LiveCap~\cite{livecap} and DeepCap~\cite{DeepCap}.
Unlike their static digital human reconstruction, the reconstruction results of our method have realistic dynamic characteristics for garments. 
%In addition, we can also animate different digital human models, as shown in \cref{fig:results}.
%
Although texture reconstruction is not the focus of this work, we present some texture reconstruction results of the reconstructed digital human in \cref{fig:results_texture} using the method of Ji et al.~\cite{relighting_ji2022}.

\subsection{Comparison and Validation}
\label{sec:comparison}
%CaPhy is the first garment simulation model for training dynamic garments with both supervised real data-driven 3D losses and unsupervised StVK physics-based losses.
We compare the proposed method with the typical data-driven digital human reconstruction method POP~\cite{POP} and the unsupervised garment simulation method SNUG~\cite{SNUG} (see \cref{fig:results}). % and the dynamic garment generation method using physical unsupervised training~\cite{SNUG}.
For a fair comparison, we apply Poisson Reconstruction to point clouds directly outputted by POP~\cite{POP} and use the same rendering conditions to generate the results. 
As SNUG~\cite{SNUG} can not obtain garment templates from real-world scans, we use our estimated template as input.

%%We construct the other compared method using a framework based on~\cite{SNUG} and the real-world fabric model described in \cref{sec:network}, which we denote as \methodone.

\begin{comment}
\begin{figure}
    \centering
    \includegraphics[width=\linewidth]{fig/results_cmp.png}
    \caption{Comparison with different garment simulation methods. From left to right: the ground-truth human scan in the test set, the results of POP~\cite{POP}, SNUG~\cite{SNUG}, \methodone, and ours. }
    \label{fig:results_cmp}
\end{figure}
    % {\includegraphics[width=0.216\linewidth]{fig/results_cmp1.png}}
    % {\includegraphics[width=0.216\linewidth]{fig/results_cmp2.png}}
    % {\includegraphics[width=0.234\linewidth]{fig/results_cmp3.png}}
    % {\includegraphics[width=0.234\linewidth]{fig/results_cmp4.png}}
\end{comment}

Limited by the size of our training set (consisting of 50 to 120 scans of a human body), the data-driven method POP~\cite{POP} has difficulties learning the dynamic characteristics of clothing for poses beyond the pose space of the training set.
%%The geometry of the generated human models is also relatively simple compared to the proposed method.
%Since POP generates the overall reconstruction of the human model, it is prone to produce incorrect or unclear semantic edges of garments.
SNUG~\cite{SNUG} and our method generate realistic clothing wrinkles.
However, since SNUG does not use 3D scanned data as constraints, the resulting clothing geometry is different from the clothing characteristics of the scanned data.

As shown in \cref{tab:quanti}, we also present a quantitative comparison of the different methods for the generated garments. 
%digital human reconstruction.
We use two sets of upper garments of human subjects for comparison.
Based on a 3D chamfer distance~\cite{POP} we evaluate the clothing geometry.
%The evaluation metric is the 3D Chamfer distance. 
Note that POP~\cite{POP} does not explicitly reconstruct the clothing, so we calculate the 3D chamfer distance from the nearest-neighbor point of the ground truth garment.
To ablate our method, we compare to our method only based on the unsupervised learning using our optimized garment physics model as described in \cref{sec:network}, which we denote as \methodone. 
Our method outperforms both the data-driven method POP and the unsupervised method SNUG~\cite{SNUG} and \methodone.
We also perform ablation experiments on the optimization of the fabric's physical parameter (denoted as \methodtwo), and find that the 3D chamfer distance remains almost unchanged before and after the optimization process.
Note that the 3D chamfer distance mainly evaluates the overall fitting accuracy between the generated garment models and the ground truth. 
%which has limited abilities to evaluate wrinkle patterns and dynamic details of garment geometry. Recent work~\cite{LaplacianFusion} demonstrates a similar experimental phenomenon: on-par (or even superior) results in terms of 3D chamfer distance do not necessarily equate to more accurate geometric reconstruction.
%
%To better evaluate the fine-detailed geometrical characteristics of the reconstructed garments,
Therefore, we also introduce a 2D perceptual metric~\cite{PerceptualLoss} 
%and structural similarity error (SSIM) 
to measure the geometric details of renderings of animated garments. 
The 2D perceptual loss is calculated through a pre-trained VGG-16 architecture~\cite{vgg16}. 
%We evaluate the rendering results without and with reconstructed textures, respectively.
%As POP does not segment the clothing region, we exclude it from the rendering comparison.
%\cref{tab:quanti} shows that our method outperforms both method \methodone and the ablation experiment on the optimization of fabric's physical parameters across all metrics.
\cref{tab:quanti} shows that our method achieves the best or equivalent results across all metrics, in comparison with POP~\cite{POP}, SNUG~\cite{SNUG}, and other ablation studies.

\begin{table}[t]
    \centering
    \begin{tabular}{l|c|c}
    \hline
    Error metric (Case 1) & 3D-CD (mm) & 2D-Perceptual \\
    \hline
	POP~\cite{POP} 		& $7.015$ 			& $0.3528$			 \\
	SNUG~\cite{SNUG} 	& $10.78$ 			& $0.4203$			 \\
	\methodone 			& $10.66$ 			& $0.4054$			 \\
	\methodtwo 			& $\mathbf{6.754}$ 	& $0.3462$		     \\
	Ours 				& $6.764$ 			& $\mathbf{0.3448}$	 \\
    \hline
    Error metric (Case 2)  & 3D-CD (mm) & 2D-Perceptual \\
    \hline
	POP~\cite{POP} 		& $8.203$ 	& $\mathbf{0.3750}$			 \\
	SNUG~\cite{SNUG} 	& $13.29$  			& $0.4749$
 \\
	\methodone 			& $12.47$			& $0.4584$
 \\
	\methodtwo 			& $7.419$ 	& $0.3906$		     \\
	Ours 				& $\mathbf{7.347}$ 	& $0.3826$	 \\
    \hline
    \end{tabular}
    \caption{Comparison of the 3D chamfer distance (3D-CD) based on the 3D reconstruction, and the 2D perceptual error based on rendering results~\cite{PerceptualLoss}.
    %\TODO{SZQ: experimenting on a fixed point subset of POP}
    %\TODO{SZQ: (1) should we eliminate POP 2D? Because 2D perceptual not only evaluates wrinkles, but also evaluates garment boundary fitting, and POP doesn't actually generate garments (2) or we should modify the evaluation text (not emphasize the "importance" of 2D-perceptual and just say "we propose 3D and 2D for evaluating...")?}
    }
    \label{tab:quanti}
\end{table}

\subsection{Limitations}
\label{sec:discussion}
For garment modeling, the proposed method utilizes clothing templates with clear topological structures to establish reasonable physical constraints for clothing.
Therefore, it may face challenges when dealing with clothing that has complex geometric structures such as pockets. 
In addition, the proposed method relies on the human template of SMPL-X~\cite{SMPL-X} to generate dynamic animations for clothing, making it difficult to handle garments that are relatively independent of the human body model, such as long dresses. 

For training part, by incorporating other constraints like 2D perceptual losses during training, we may achieve results with higher fidelity in future work. Also, when performing physical parameter optimization, we fix some basic parameters (e.g. density) to avoid parameter degeneracy, future works may explore a more decent method to perform better physical optimization. In addition, with our collision fine-tune step, the garment-garment intersection problem is not fully solved and needs further improvement.
%Future research will aim to establish a clothing simulation and digital human reconstruction system that is suitable for physical model constraints, capable of supporting a wider range of clothing styles and enabling a more flexible dynamic generation of clothing.

\section{Conclusion}
\label{sec:conclusion}
We introduced CaPhy, a digital human avatar reconstruction method that is based on an optimizable physics model to learn a neural garment deformation model which extrapolates to novel poses not seen in the input 3D scans.
Specifically, we combine unsupervised physics-based constraints and 3D supervision to reproduce the physical characteristics of the real garments from observations.
We demonstrate that this method, allows us to reconstruct an avatar with clothing that extrapolates to novel poses with realistic product of wrinkles.
%This approach eliminates the gap between physical constraints and real-world data and addresses the challenge of generalizing the dynamic characteristics of clothing to poses beyond the dataset when relying solely on data-driven methods.
%Second, we introduce an optimizable cloth physical model that improves the generation of realistic digital humans by optimizing the physical parameters of clothing.
We believe that CaPhy is a stepping stone towards generalizable avatar animation that combines physics with sparse observations.
%Our method provides inspiration for future research in dynamic clothing generation through the combination of unsupervised and supervised training.
%In addition, it paves the way for the development of physical models of clothing based on scanned data.

\textbf{Acknowledgements.} This paper is supported by National Key R\&D Program of China (2022YFF0902200), the NSFC project No.62125107 and No.61827805.

{\small
\bibliographystyle{ieee_fullname}
\bibliography{refs}
}

\end{document}